\documentclass[letterpaper]{article} 
\usepackage{aaai25}  
\usepackage{times}  
\usepackage{helvet}  
\usepackage{courier}  
\usepackage[hyphens]{url}  
\usepackage{graphicx} 
\usepackage{natbib}  
\usepackage{caption} 
\usepackage{algorithm}
\usepackage{algorithmic}
\usepackage{threeparttable}
\usepackage{booktabs}
\usepackage{multirow}
\usepackage{amssymb} 
\usepackage{mathrsfs}
\usepackage{amsmath}
\usepackage{amsmath}
\usepackage{xcolor}

\usepackage{newfloat}
\usepackage{listings}
\DeclareCaptionStyle{ruled}{labelfont=normalfont,labelsep=colon,strut=off} 
\floatstyle{ruled}
\newfloat{listing}{tb}{lst}{}
\floatname{listing}{Listing}
\title{Hypergraph Attacks via Injecting Homogeneous Nodes into Elite Hyperedges}
\author {
    Meixia He\textsuperscript{\rm 1},
    Peican Zhu\textsuperscript{\rm 1}\thanks{P. Zhu, K. Tang, Y. Guo are joint corresponding authors.},
    Keke Tang\textsuperscript{\rm 2,3}\footnotemark[1],
    Yangming Guo\textsuperscript{\rm 4}\footnotemark[1]
}
\affiliations {
    \textsuperscript{\rm 1}School of Artificial Intelligence, Optics and Electronics (iOPEN), Northwestern Polytechnical University\\
    \textsuperscript{\rm 2}Cyberspace Institute of Advanced Technology, Guangzhou University\\
    \textsuperscript{\rm 3}Huangpu Research School, Guangzhou University\\    
    \textsuperscript{\rm 4}School of Cybersecurity, Northwestern Polytechnical University\\
    meixia\_he@mail.nwpu.edu.cn, ericcan@nwpu.edu.cn, tangbohutbh@gmail.com, yangming\_g@nwpu.edu.cn
}

\usepackage{bibentry}

\begin{document}

\maketitle

\begin{abstract}
Recent studies have shown that Hypergraph Neural Networks (HGNNs) are vulnerable to adversarial attacks. Existing approaches focus on hypergraph modification attacks guided by gradients, overlooking node spanning in the hypergraph and the group identity of hyperedges, thereby resulting in limited attack performance and detectable attacks. In this manuscript, we present a novel framework, i.e., Hypergraph \textbf{Attack}s via \textbf{I}njecting Homogeneous Nodes into \textbf{E}lite Hyperedges (IE-Attack), to tackle these challenges. Initially, utilizing the node spanning in the hypergraph, we propose the elite hyperedges sampler to identify hyperedges to be injected. Subsequently, a node generator utilizing Kernel Density Estimation (KDE) is proposed to generate the homogeneous node with the group identity of hyperedges. Finally, by injecting the homogeneous node into elite hyperedges, IE-Attack improves the attack performance and enhances the imperceptibility of attacks. Extensive experiments are conducted on five authentic datasets to validate the effectiveness of IE-Attack and the corresponding superiority to state-of-the-art methods. 
\end{abstract}  
%
\section{Introduction}
Graph Neural Networks (GNNs) 
\cite{kipf2016semi,velivckovic2017graph,cheng2024heuristic,liu2023graph,dong2023generalized,wang2023augmenting,cheng2024gin} capture intricate relationships and patterns within graph data, facilitating tasks like node classification, recommendation systems and source detection, etc. Nevertheless, with the increasing complexity and diversity of real-world networks, attention has shifted towards higher-order networks like hypergraphs and simplicial complexes \cite{benson2016higher,battiston2021physics,10771713}. Subsequently, Hypergraph Neural Networks (HGNNs) \cite{feng2019hypergraph,bai2021hypergraph} are proposed to extract higher-order features from hypergraphs, significantly improving the efficiency of downstream graph-related tasks. Despite its success, GNNs and HGNNs have been shown to be vulnerable to adversarial attacks 
 \cite{zugner2018adversarial,bojchevski2019adversarial}, which has attracted increasing research interest.   
\begin{figure}[!htbp]
  \centering 
  \includegraphics[width=0.98\linewidth]{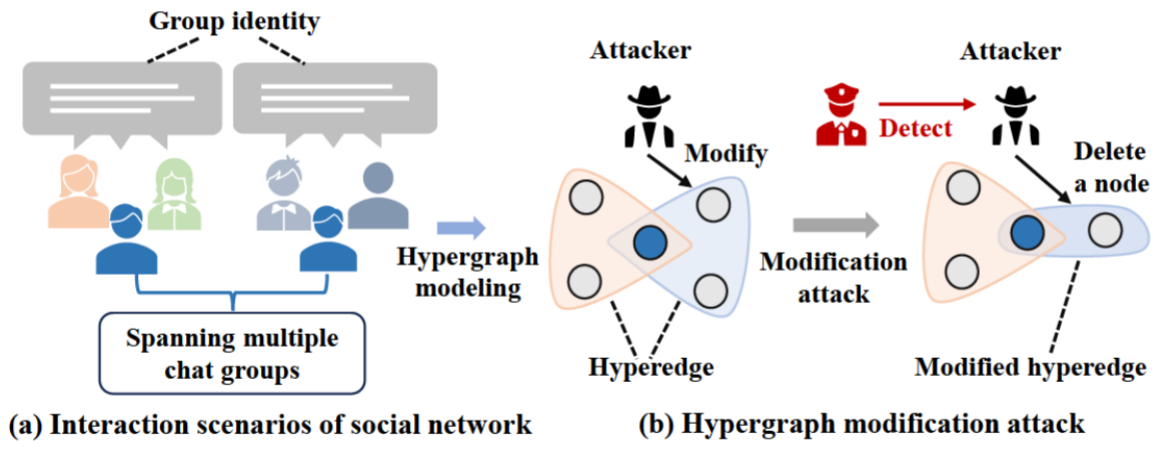}
  \caption{An illustration of the attack scenarios of hypergraph. (a) The same user spans multiple chat groups and chat groups exhibit group identity in social network. (b) Current hypergraph attack methods only involve hypergraph modification attacks and are easily detectable.}
  \label{motivation}
\end{figure}

Adversarial attacks on GNNs are divided into modification attack and injection attack according to attack techniques \cite{wei2020adversarial,sun2022adversarial}. In modification attack \cite{chang2020restricted,jin2023local}, attackers aim to affect the performance of GNNs by modifying the attributes or connections of nodes and edges. Conversely, due to the high authority required by modification attack, the injection attack has received more attention. For instance, in social networks, injection attacks involve create new fake accounts without the need to compromise existing accounts, thus requiring relatively low permissions. Injection attacks \cite{sun2019node,zou2021tdgia,zhu2024node} inject new nodes or edges into the graph, introducing malicious information to degrade the performance of GNNs. Nevertheless, researches on adversarial attacks against HGNNs are limited to the hypergraph modification attack, including HyperAttack \cite{hu2023hyperattack} and MGHGA \cite{chen2023momentum}. These methods do not further consider the phenomenon of node spanning in the hypergraph and the group identity of hyperedges, resulting in poor attack performance and easily detectable attacks, as illustrated in Figure \ref{motivation}. 

By modeling higher-order relationships in the network, hypergraphs exhibit richer graph structural features, leading to more limitations on hypergraph attacks. First, modifying hypergraphs requires high authority, limiting the applicability of hypergraph modification attacks. Second, the existence of node spanning in hypergraphs \cite{battiston2021physics} implies that when the same node appears in multiple hyperedges, hyperedges with substantial influence can be discerned by the frequency of node occurrences across various hyperedges. For instance, in social networks, when a user presents in multiple chat groups, it suggests that the user can extensively disseminate information through these groups within the network. However, current methods select modified hyperedges by calculating gradients, which cannot maximize the destruction of the feature aggregation of HGNNs, resulting in poor attack performance. Moreover, hyperedges are viewed as groups and exhibit group identity according to social psychology research \cite{spears2021social}. Current hypergraph modification attacks do not consider the group identity of hyperedges when adding or removing known nodes in hyperedges, making the attacks easily detectable.
  
To address these challenges, we propose the Hypergraph \textbf{Attack}s via \textbf{I}njecting Homogeneous Nodes into \textbf{E}lite Hyperedges (IE-Attack). Firstly, due to the high authority required by hypergraph modification, we present the node injection attacks on hypergraphs. Secondly, inspired by the elite group in social psychology \cite{shayegh2022social,howard2000social}, we propose an elite hyperedges sampler to identify elite hyperedges with significant influence by utilizing the node spanning phenomenon in the hypergraph. Additionally, we develop a node generator based on Kernel Density Estimation (KDE) to generate the homogeneous node, so that the elite hyperedges injected into the homogeneous node still exhibit group identity. By injecting the homogeneous node into elite hyperedges, IE-Attack maximizes malicious information propagation to obtain excellent attack results and enhances the imperceptibility of attacks. We validate the effectiveness of our approach on five publicly available datasets. Extensive experimental results demonstrate that IE-Attack exhibits excellent attack performance on HGNNs, outperforming state-of-the-art node injection attack methods. 

Overall, our contributions are summarized as:
\begin{itemize}
    \item We are the first to analyze the problem of node injection attack in hypergraphs and to propose a methodology to address this challenge.
    \item We propose a novel attack method against HGNNs for the Hypergraph Attacks via Injecting Homogeneous Nodes into Elite Hyperedges.  
    \item We demonstrate the effectiveness of our proposed
    method over baseline approaches through extensive experimental validation.
\end{itemize}

\section{Related Works}
\subsubsection{Graph Injection Attack}
Deep neural networks \cite{tang2022decision} are known to be vulnerable to adversarial attacks, and this susceptibility has been extensively studied across various domains, including images \cite{zhu2024improving,li2023hept}, point clouds \cite{tang2022rethinking}, and graphs \cite{wei2020adversarial}.
In this paper, we focus on graph injection attacks. NIPA \cite{sun2019node} introduced a novel node injection approach to poison the graph structure, which was based on reinforcement learning. Likewise, AFGSM \cite{wang2020scalable} addressed a more practical attack scenario, allowing adversaries to inject malicious nodes into the graph without manipulating the existing graph structure. Conversely, both NIPA and AFGSM were developed in a poisoning environment, requiring retraining of the defense models after each attack. TDGIA \cite{zou2021tdgia} followed the evasion attack setting of KDDCUP 2020, where different attacks were evaluated based on the same set of models and weights. In contrast to previous attacks in a white-box setting, G$^{2}$A2C \cite{ju2023let} and \cite{ijcai2024p0063} proposed node injection attacks in a black-box setting. Subsequently, considering that previous studies involved injecting multiple nodes for attacks, G-NIA \cite{tao2021single} introduced the extreme scenario of single node injection. G-NIA\_CANA \cite{tao2023adversarial} incorporated a generator-discriminator structure to enhance the imperceptibility of attacks based on G-NIA.

Different from node injection attack in ordinary graphs, hyperedges involve higher-order relationships among multiple nodes in hypergraphs, rendering graph injection attack methods unsuitable for hypergraph structures. 
\subsubsection{Hypergraph Neural Networks} 
To encapsulate more intricate higher-order interaction details within graph datasets, advanced structures like hypergraphs, represented as \(\mathcal{G}=(\mathcal{V},\mathcal{E})\), have been identified and developed \cite{jin2019robust,antelmi2023survey}. Subsequently, HGNNs \cite{feng2019hypergraph} were introduced, where the feature aggregation process involved node-hyperedge-node interactions. This enabled HGNNs to extract more comprehensive feature information compared to GNNs. Approaches such as hypergraph convolution and hypergraph attention \cite{bai2021hypergraph,yadati2019hypergcn,zhang2019hyper} delved into refining the feature aggregation mechanism within hypergraphs, demonstrating notable effectiveness across graph-related tasks like node classification and link prediction. Additionally, a range of related HGNNs provided varied perspectives and find applications in fields such as recommendation systems and biological information networks \cite{zeng2023multi,wang2018dual,yu2023basket}.

While the security of HGNNs warrants significant attention, current attacks on HGNNs are solely focused on hypergraph modifications attacks \cite{hu2023hyperattack,chen2023momentum}, overlooking the node spanning phenomenon in the hypergraph and the group identity within hyperedge. 
\begin{figure*}[!htbp]
  \centering
  \includegraphics[width=0.98\linewidth]{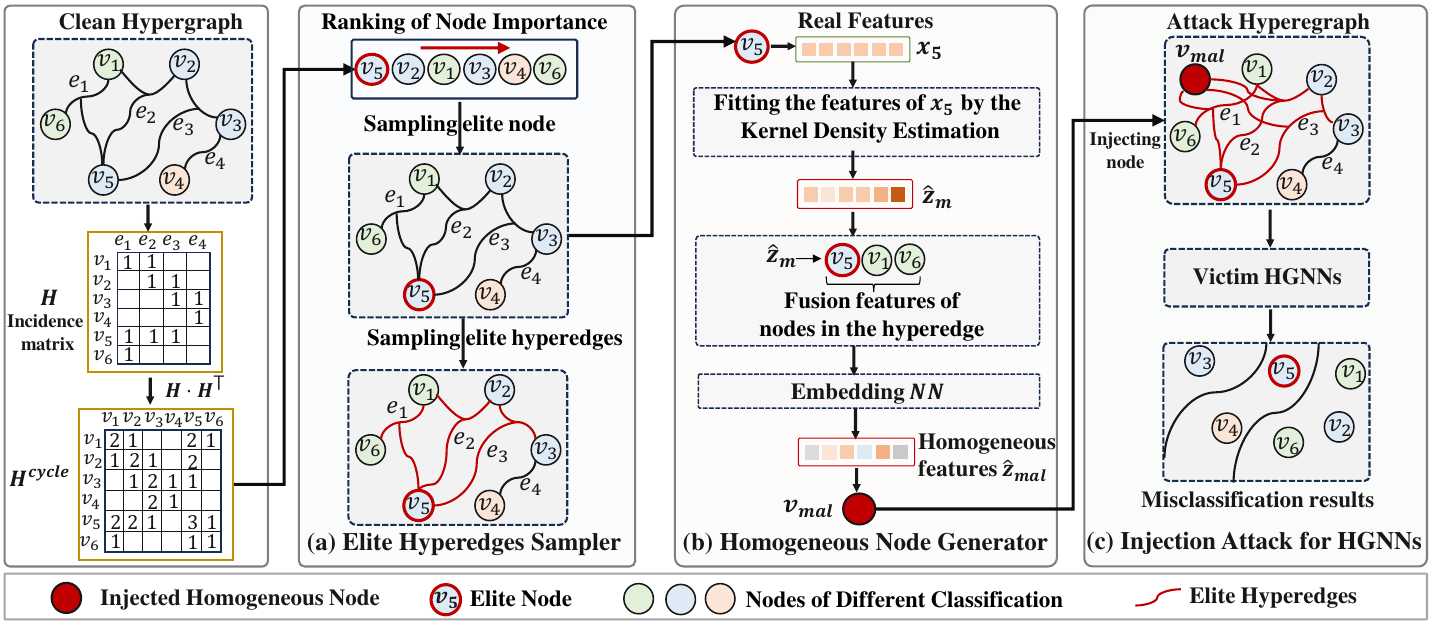}
  \caption{Framework of IE-Attack. (a) Elite Hyperedges Sampler obtains elite hyperedges. (b) Homogeneous Node Generator 
 generates the homogeneous node. (c) IE-Attack attacks HGNNs by injecting the homogeneous node into elite hyperedges.}
  \label{frame}
\end{figure*}
\section{Preliminary and Problem Statement}
\subsubsection{Graph Neural Networks}
The graph $G = (V, E)$, where $V$ represents the set of nodes and $E$ represents pairwise edges. The node features are denoted by $X \in \mathbb{R}^{|V|\times|F|}$, where $F$ signifies the feature dimension of $X$. The adjacency matrix is represented as $A\in \mathbb{R}^{|V|\times|V|}$, with $A_{ij}$ being 1 if there exists a connection between node $v_{i}$ and $v_{j}$, and 0 otherwise. The GNNs involve a graph convolution process, where the feature aggregation is expressed as:
\begin{align}\label{GNN_eq}
X^{l+1}=\sigma(D^{-\frac{1}{2}}(A+I)D^{-\frac{1}{2}}X^{l}W^{l}),
\end{align}
where $D$ denotes the degree matrix, $I$ stands for the identity matrix, and $W^{l}$ represents the weight matrix of the $l$-th layer in GNNs. The symbol $\sigma$ signifies the activation function.
\subsubsection{Hypergraph Neural Networks}
In contrast to graph $G = (V, E)$, hypergraph defined as $\mathcal{G}=(\mathcal{V},\mathcal{E})$, where $\mathcal{V}$ represents the set of nodes and $\mathcal{E}$ denotes hyperedges sets. The feature matrix $\mathcal{X} \in \mathbb{R}^{|\mathcal{V}|\times|\mathcal{F}|}$ is the node features, with $\mathcal{F}$ denoting the feature dimension of $\mathcal{X}$. Hyperedges consist of sets of nodes comprising two or more nodes, such as the hyperedge $e_{1}=\{v_{1},v_{5},v_{6}\}$ in Figure \ref{frame}. The incidence matrix $H$, which delineates the relationships between nodes and hyperedges, is represented as follows:
\begin{align}\label{H}
H = \left\{
\begin{aligned}
  &H_{ij}=1,~~~~ if~~~v_{i} \in e_{j}, \\ 
  &H_{ij}=0,~~~~ if~~~v_{i} \notin e_{j}. \\
\end{aligned}
\right.
\end{align}
Then, the feature aggregation of HGNNs is defined as:
\begin{align}\label{HGNN_eq}
\mathcal{X}^{l^{h}+1}=\sigma(D_{\mathcal{V}}^{-\frac{1}{2}}H\mathcal{W}D_{\mathcal{E}}^{-1}H^{\mathsf{T}}D_{\mathcal{V}}^{-\frac{1}{2}}\mathcal{X}^{l^{h}}\Theta^{l^{h}}),
\end{align}
where $D_{\mathcal{V}}$ signifies the degree matrix of nodes, $D_{\mathcal{E}}$ represents the degree matrix of hyperedges, and $\mathcal{W}$ denotes the weight associated with hyperedges. The parameters for training $\Theta^{l^{h}}$ correspond to the $l^{h}$-th layer in HGNNs.
\subsubsection{Main Challenges of Node Injection Attack in HGNNs}
In the ordinary graph $G=(V,E)$, the connections between nodes are pairwise. When discussing node injection attacks on GNNs, the focus lies in introducing new nodes into the node set within the graph. The key steps are as follows: (1) the attacker selects target nodes for the injection of new nodes; (2) the attacker generates new nodes and inserts them into the chosen target nodes, thereby establishing new edges.

Conversely, due to the node spanning phenomenon in the hypergraph, the information propagation capacity of nodes across different hyperedges varies. Additionally, hyperedges exhibit group identity, leading to vigilance towards injected nodes and making attacks detectable. Consequently, the attack against HGNNs have two main challenges: (1) proposing a method that leverages the node spanning phenomenon to identify elite hyperedges for node injection attacks, maximizing the spread of malicious information; (2) Considering the group identity of hyperedges, a node generator should be proposed to enhance the imperceptibility of attacks. 
\section{Methodology}
\subsubsection{Elite Hyperedges Sampler} 
The phenomenon of node spanning in the hypergraph indicates varying abilities of nodes to propagate information within hyperedges, providing us with insights into selecting hyperedges for node injection.
Initially, we analyze the feature aggregation process of HGNNs, $\mathcal{X}^{l^{h}+1}=\sigma(D_{\mathcal{V}}^{-\frac{1}{2}}H\mathcal{W}D_{\mathcal{E}}^{-1}H^{\mathsf{T}}D_{\mathcal{V}}^{-\frac{1}{2}}\mathcal{X}^{l^{h}}\Theta^{l^{h}})$. The purpose of $D_{\mathcal{V}}$ and $D_{\mathcal{E}}$ is to normalize the matrix $H$, further simplified to $\mathcal{X}^{l^{h}+1}=H\mathcal{W}H^{\mathsf{T}}\mathcal{X}^{l^{h}}\Theta^{l^{h}}$. Considering the $\mathcal{W}$ as the identity matrix, $\mathcal{X}^{l^{h}+1}=HH^{\mathsf{T}}\mathcal{X}^{l^{h}}\Theta^{l^{h}}$ is obtained. Subsequently, the feature aggregation process can be simplified and represented as follows:
\begin{align}\label{feature}
\mathcal{X}^{l^{h}+1}=H \cdot H^{\mathsf{T}} \cdot \mathcal{X}^{l^{h}}.
\end{align}
During node feature updates, we first aggregate hyperedge features using $H^{\mathsf{T}} \cdot \mathcal{X}^{l^{h}}$, followed by updating node features using $H \cdot H^{\mathsf{T}} \cdot \mathcal{X}^{l^{h}}$.
The $H \cdot H^{\mathsf{T}}$ signifies the weight adjustments post feature updates, reflecting the frequency of shared hyperedges between pairs of nodes. Consequently, a higher correlation coefficient arises from the increased frequency of shared hyperedges between two nodes, resulting in more enriched features during the aggregation process. Given that nodes exist across multiple hyperedges, does this suggest that these nodes offer substantial feature information during the aggregation process? 

Examining the cyclic structure of the network tackles this challenge and introduces the notion of ``cycle ratio" quantifying a node's involvement in the shortest cycles of other nodes \cite{fan2021characterizing}. Inspired by this, if we consider hyperedge as cycles, then $H^{cycle}=H\cdot H^{\mathsf{T}}$ symbolizes the ``cyclenumber matrix" within the hypergraph. The sum of the participation rates of node $v_{i}$ in the hyperedge involving itself and other nodes is the ``cycle ratio", which is:
\begin{align}\label{elite}
p_{v_{i}}=\sum\nolimits_{j=1}^{\mu_{\mathcal{E}}}\frac{H_{ij}^{cycle}}{H_{jj}^{cycle}},
\end{align}
where $H_{ij}^{cycle}$ denotes the count of hyperedges in which both nodes $v_i$ and $v_j$ are participants, while $H_{jj}^{cycle}$ represents the number of hyperedges involving node $v_{j}$. $\mu_{\mathcal{E}}$ denotes the number of hyperedges. $p_{v_{i}}$ reflects the importance of node $v_{i}$ within the hypergraph. This novel idea suggests that the more involved an individual is within a community (sharing hyperedges with neighbors) and the wider range of social roles they assume (including hyperedges they are associated with), the more significant their role becomes. Subsequently, we calculate the ``cycle ratio" of all nodes in the hypergraph through Eq. (\ref{elite}) and sorting them, the most important elite node $v_{elite}$ is obtained. The hyperedge that includes elite node $v_{elite}$ is recognized as elite hyperedges $\mathcal{E}_{elite}$.

\subsubsection{Homogeneous Node Generator}
To generate homogeneous nodes with features resembling the elite node, we employ Kernel Density Estimation function $\hat{f}_{K_{de}}(x)$ \cite{wkeglarczyk2018kernel}, defined as:
\begin{align}\label{KDE}
\hat{f}_{K_{de}}(x) = \frac{1}{\mathcal{F}} \sum\nolimits_{i=1}^{\mathcal{F}} K_{de}(x - x_{i}).
\end{align}

Here, $\mathcal{F}$ denotes the sample size, which is equivalent to the dimensionality of the features. $x_{i}$ represents $i$-th data point of features vector $x$. $K_{de}$ refers to the kernel function, utilized to assess the density contribution in the vicinity of data points. By placing kernel functions around data points and blending them with weights, a smoothed density estimation is achieved over the entire spatial domain. 

Initially, we utilize the KDE function $\hat{f}_{K_{de}}(x)$ to fit features of the elite node $v_{elite}$. The fitting process is as follows:
\begin{align}\label{nihe}
\Phi_{elite}=\hat{f}_{K_{de}}(z_{elite}),
\end{align}
where $z_{elite}$ denotes the feature of $v_{elite}$. $\Phi_{elite}$ is the probability density function obtained by fitting the features $z_{elite}$. 

Introducing an excessive number of homogeneous nodes elevate the dimensionality of the $H$ and escalate the computational complexity of the HGNNs. Consequently, $\Phi_{elite}$ are sampled to obtain preliminary features $\widehat{z}_{m}$ of single homogeneous node $v_{mal}$ with a similar distribution of features as the elite node $v_{elite}$. The dimension $\mathcal{F}$ of $\widehat{z}_{m}$ is the same as features of $v_{elite}$. This process can be formalized as:
\begin{align}\label{vk}
\widehat{z}_{m}=\Phi_{elite}.sample(1\times\mathcal{F}).
\end{align}

While the $v_{mal}$ exhibits group identity of the elite hyperedge similarity to $v_{elite}$, features of nodes other than $v_{elite}$ within the hyperedge are also essential. Given that $v_{elite}$ exists in multiple elite hyperedges, selecting all elite hyperedges will increase complexity. Therefore, we randomly select an elite hyperedge $e_{elite}=\{v_{1},\ldots,v_{t}\}$ and further optimize the generation of $\widehat{z}_{m}$ using the feature information from other nodes in hyperedge $e_{elite}$, represented as follows:
\begin{align}\label{mean}
&\widehat{z}_{ms}=\sigma(((\widehat{z}_{m}\otimes \Theta^{1})\otimes \Theta^{2})\cdots\otimes \Theta^{l^{h}}),\\
&\widehat{z}_{msd}=r_{elite}\oplus r_{mean}\oplus \widehat{z}_{ms},\\
&\widehat{z}_{mal}=NN(\widehat{z}_{msd}).
\end{align}
Herein, $r_{mean}=\frac{r_{1}+\cdots+r_{t}}{t-1}$, $\{r_{1},\ldots,r_{t}\}$ denote the node embedding acquired for nodes $\{v_{1},\ldots,v_{t}\}$ via the surrogate model HGNNs. $r_{elite}$ is node embedding of $v_{elite}$ in $e_{elite}$. $t$ signifies the size of the elite hyperedge $e_{elite}$. $\{\Theta^{1},\Theta^{2},\ldots,\Theta^{l^{h}}\}$ represent the inter-layer weight parameters trained within the surrogate model HGNNs. The $\otimes$ denotes the element-wise multiplication operation between vectors, while $\oplus$ signifies the concatenation operation between vectors. $\widehat{z}_{ms}$ denotes the homogeneous feature vector further computed using the trained weights in HGNNs. Subsequently, the homogeneous feature vector $\widehat{z}_{msd}$ undergoes training through a linear neural network layer $NN$, where $NN$ is expressed as $Y=W_{NN}\widehat{z}_{msd}+b_{offset}$. Here, $W_{NN}$ denotes the weight parameters of $NN$, and $b_{offset}$ represents the bias term. Consequently, we obtain the homogeneous node $v_{mal}$ with homogeneous feature $\widehat{z}_{mal}$.

\subsubsection{Injection Attack for HGNNs}
To achieve excellent attack performance with a lower attack cost, we inject single homogeneous node $v_{mal}$ into $\mathcal{E}_{elite}$. The generated homogeneous node $v_{mal}$ serves as an attacker injected into elite hyperedges $\mathcal{E}_{elite}$, thereby propagating malicious information during feature aggregation of HGNNs and increasing the imperceptibility of attacks.  

Given the hyperedge incidence matrix of the original hypergraph $\mathcal{G}$ as $H^{\mu_{\mathcal{V}} \times \mu_{\mathcal{E}}}$, where $\mu_{\mathcal{V}}$ is the number of nodes and $\mu_{\mathcal{E}}$ is the number of hyperedges. The single homogeneous node $v_{mal}$ is injected into elite hyperedges $\mathcal{E}_{elite}$, resulting in the attacked hypergraph $\widehat{\mathcal{G}}$. The hyperedge incidence matrix of $\widehat{\mathcal{G}}$ is updated to $\widehat{H}^{(\mu_{\mathcal{V}}+1) \times \mu_{\mathcal{E}}}$.
 
For instance, if $\mathcal{E}_{elite} = \{e_{elite1},\ldots,e_{elitej}\}$ and $e_{elitej} = \{v_{1},v_{5},v_{i}\}$, injecting the homogeneous node $v_{mal}$ into $e_{elitej}$, results in the attacked hyperedge $\widehat{e}_{elitej}=\{v_{1},v_{5},v_{i},v_{mal}\}$. Consequently, the value at the corresponding position $\{(\mu_{\mathcal{V}}+1),j\}$ in the $\widehat{H}^{(\mu_{\mathcal{V}}+1) \times \mu_{\mathcal{E}}}$ matrix is set to 1. $(\mu_{\mathcal{V}}+1)$ is the position of $v_{mal}$ in the node set $\mathcal{V}$. $j$ denotes the $j$-th hyperedge of the elite hyperedge $e_{elitej}$ in the hyperedges set $\mathcal{E}$. 

The attacked $\widehat{H}$ resulting from injecting the homogeneous node $v_{mal}$ into elite hyperedges $\mathcal{E}_{elite}$ is described as:
\begin{align}\label{Her}
\widehat{H} = \left\{
\begin{aligned}
  &\widehat{H}_{ij}=1,~~~~ if~~~v_{i} \in e_{j}, \\ 
  &\widehat{H}_{ij}=0,~~~~ if~~~v_{i} \notin e_{j}, \\
  &\widehat{H}_{(\mu_{\mathcal{V}}+1)j}=1,~~~~ if~~~v_{mal} \in e_{elitej}. \\
\end{aligned}
\right.
\end{align}

It is evident that the node dimension of $\widehat{H}$ has increased by one, while the hyperedge dimension remains unchanged. Additionally, the feature matrix $\mathcal{X}$ has been augmented with the injected homogeneous node feature $\widehat{z}_{mal}$.
Therefore, we obtain the input data for adversarial attacks on HGNNs, comprising the attacked incidence matrix $\widehat{H}$ and the perturbed feature attributes $\widehat{\mathcal{X}}=\{\mathcal{X},\widehat{z}_{mal}\}$.

HGNNs aggregate the malicious feature information in the attacked hypergraph, leading to a degradation in HGNNs. $\widehat{\mathcal{Z}}$ is the output of HGNNs after being attacked:
\begin{align}\label{miss}
\widehat{\mathcal{Z}}=\sigma(\widehat{D}_{\mathcal{V}}^{-\frac{1}{2}}\widehat{H}\widehat{\mathcal{W}}\widehat{D}_{\mathcal{E}}^{-1}\widehat{H}^{\mathsf{T}}\widehat{D}_{\mathcal{V}}^{-\frac{1}{2}}\widehat{\mathcal{X}}^{l^{h}}\Theta^{l^{h}}).
\end{align}
\subsubsection{Optimization}
Finally, we aim to enhance the attack effectiveness by training the IE-Attack, thereby improving the \textit{Misclassification rate} of HGNNs. Consequently, the objective of model training is to minimize the discrepancy between predicted scores of correct labels and the predicted scores of the targeted incorrect labels, being depicted as: 
\begin{align}\label{loss}
\min_{\widehat{\mathcal{G}}} \mathcal{L}_{atk}=\sum\nolimits_{q\in \mathcal{V}_{train}}\max_{z \neq y_{q}}(0,\widehat{\mathcal{Z}}_{q,y_{q}}-\widehat{\mathcal{Z}}_{q,z})+ \nonumber \\
\parallel \widehat{z}_{mal}-z_{elite}\parallel,
\end{align}
where $q$ denotes the index of training samples in the training set $\mathcal{V}_{train}$, $y_{q}$ indicates the label of sample $q$, $\widehat{\mathcal{Z}}$ represents the classification prediction results of HGNNs on the attacked hypergraph $\widehat{\mathcal{G}}$, and $z$ is the label of sample $q$ obtained by the surrogate model HGNNs. The ReLU function $\max(0, \cdot)$ is employed to ensure the non-negativity of the loss function. The term $\|\widehat{z}_{mal} - z_{elite}\|$ represents the distance between the embeddings of the elite node $v_{elite}$ and the generated homogeneous node $v_{mal}$. The purpose is to make the generated homogeneous node $v_{mal}$ closer to the elite node $v_{elite}$ with the group identity of the elite hyperedges, thereby improving the imperceptibility of attacks. The training process of the IE-Attack is guided by the attack loss $\mathcal{L}_{atk}$, optimizing iteratively with the gradient descent method until convergence. 
\section{Experiments}
\subsection{Experimental Setting}
\begin{table*}[!htbp]\small 
\centering
\setlength{\tabcolsep}{3pt}
    \begin{tabular}{l|l|c|cc|cc|ccl}
        \toprule
        Datasets&Hyper-model&Clean&GIA-R &DICE &FGA &IGA & G-NIA&  G-NIA$^{*}$& \multicolumn{1}{c}{IE-Attack}\\
        \hline
        \multirow{3}{*}{Cora}& Hyper-KNN
        &25.24$\pm$2.69&25.60$\pm$2.71&25.65$\pm$2.71&25.86$\pm$2.71&25.21$\pm$2.69&52.09$\pm$3.09&34.50$\pm$2.94&\textbf{55.92}$\pm$\textbf{3.02} \\
        & Hyper-HOR &25.35$\pm$2.70&25.48$\pm$2.70&25.68$\pm$2.71&25.68$\pm$2.71&25.53$\pm$2.70&37.51$\pm$2.99&29.57$\pm$2.83&\textbf{46.37}$\pm$\textbf{3.09} \\
        & Hyper-$L1$ &35.30$\pm$2.96&35.77$\pm$2.97&36.48$\pm$2.98&36.16$\pm$2.98&35.75$\pm$2.97&42.15$\pm$3.05&39.60$\pm$2.30&\textbf{55.60}$\pm$\textbf{3.08} \\
        \hline
        \multirow{3}{*}{Citeseer}& Hyper-KNN&42.23$\pm$3.07&42.26$\pm$3.06&42.32$\pm$3.06&42.55$\pm$3.06&42.96$\pm$3.07&52.93$\pm$5.62&56.93$\pm$3.06&\textbf{72.00}$\pm$\textbf{2.78} \\
        & Hyper-HOR&45.51$\pm$3.09&45.31$\pm$3.09&45.57$\pm$3.09&45.33$\pm$3.09&45.35$\pm$3.09&45.30$\pm$5.64&49.27$\pm$3.10&\textbf{53.20}$\pm$\textbf{3.09} \\
        & Hyper-$L1$ &43.20$\pm$3.07&43.71$\pm$3.08&43.14$\pm$3.07&44.59$\pm$3.08&43.60$\pm$3.07&59.53$\pm$4.55&49.67$\pm$5.66&\textbf{70.61}$\pm$\textbf{2.82} \\
        \hline
        
        \multirow{3}{*}{Pubmed} & Hyper-KNN&18.05$\pm$1.06&18.15$\pm$1.07&18.17$\pm$1.07&18.34$\pm$1.07&18.16$\pm$1.07&20.64$\pm$1.03&20.15$\pm$1.06&\textbf{22.42}$\pm$\textbf{1.15} \\
        &Hyper-HOR&16.91$\pm$1.04&16.66$\pm$1.03&16.79$\pm$1.04&16.77$\pm$1.04&16.74$\pm$1.04&23.18$\pm$1.45&23.94$\pm$1.37&\textbf{26.13}$\pm$\textbf{1.22} \\
        & Hyper-$L1$ &22.80$\pm$1.24&22.81$\pm$1.16&22.71$\pm$1.16&22.77$\pm$1.16&22.92$\pm$1.16&24.06$\pm$1.06&24.85$\pm$1.07&\textbf{27.02}$\pm$\textbf{1.18} \\
        \hline
        \multirow{3}{*}{Chameleon} & Hyper-KNN&40.98$\pm$4.32&41.24$\pm$4.32&40.50$\pm$4.31&41.10$\pm$4.31&41.04$\pm$4.32&46.64$\pm$4.36&43.60$\pm$4.16&\textbf{47.98}$\pm$\textbf{4.38} \\
        &Hyper-HOR&55.56$\pm$4.35&57.16$\pm$4.34&56.62$\pm$4.35&56.42$\pm$4.34&56.54$\pm$4.35&58.86$\pm$4.32&59.40$\pm$4.31&\textbf{70.26}$\pm$\textbf{3.96} \\
        & Hyper-$L1$ &74.20$\pm$3.84&74.92$\pm$3.08&75.24$\pm$3.78&75.04$\pm$3.79&75.50$\pm$3.77&75.80$\pm$2.67&75.20$\pm$3.79&\textbf{79.00}$\pm$\textbf{3.57}\\
        \hline
        \multirow{3}{*}{Lastfm} & Hyper-KNN&63.85$\pm$1.71&66.52$\pm$1.69&66.58$\pm$1.69&66.47$\pm$1.69&66.26$\pm$1.69&65.26$\pm$4.18&65.96$\pm$1.67&\textbf{67.81}$\pm$\textbf{1.67} \\
        &Hyper-HOR&56.71$\pm$1.76&56.90$\pm$1.78&56.38$\pm$1.77&56.95$\pm$1.77&56.80$\pm$1.77&59.58$\pm$4.30&62.53$\pm$1.57&\textbf{64.57}$\pm$\textbf{1.71} \\
        & Hyper-$L1$&70.00$\pm$1.61&70.44$\pm$1.26&70.42$\pm$1.62&70.27$\pm$1.62&70.43$\pm$1.67&70.40$\pm$3.01&69.77$\pm$1.44&\textbf{72.87}$\pm$\textbf{1.16} \\
        \bottomrule
    \end{tabular}
    \caption{Comparison of \textit{Misclassification rate} (\%) of IE-Attack and baselines. The results are averaged over 10 runs and the half width of the 95\% confidence interval. The best results are highlighted in bold.}
    \label{performance_all}
\end{table*}

\begin{table*}[htbp]\small
\centering
\setlength{\tabcolsep}{1.6pt}
    \begin{tabular}{l|c|ccccccc|ccccccc}
        \toprule
        \multirow{2}{*}{Dataset} &\multirow{2}{*}{Hyper-model}&\multicolumn{7}{c|}{PCA} &\multicolumn{7}{c}{HBOS}\\
        \cline{3-16}
        & &GIA-R &DICE &FGA &IGA&G-NIA&G-NIA$^{*}$&IE-Attack&GIA-R &DICE &FGA &IGA&G-NIA&G-NIA$^{*}$&IE-Attack\\
        \hline
        \multirow{1}{*}{Cora}& Hyper-KNN&25.40&26.10&26.00&25.90&45.40&25.30&\textbf{49.50}&26.70&26.70&26.90&27.10&45.40&26,40&\textbf{57.80}\\
        \hline
        \multirow{1}{*}{Citeseer}& Hyper-KNN&40.70&40.00&40.90&40.20&42.60&41.10&\textbf{79.20}&43.00&43.70&42.70&42.80&43.10&44.30&\textbf{78.40}\\
        \hline
        \multirow{1}{*}{Pubmed} &Hyper-KNN&17.92&17.80&17.96&17.80&19.05&19.23&\textbf{20.02}&18.04&18.06&18.50&18.64&20.14&20.21&\textbf{22.62}\\
        \hline
        \multirow{1}{*}{Chameleon} & Hyper-KNN&39.60&39.60&39.60&39.60&44.60&44.20&\textbf{45.60}&42.00&42.20&41.40&41.80&46.20&39.40&\textbf{46.57}\\
        \hline
        \multirow{1}{*}{Lastfm} &Hyper-KNN&64.76&65.00&64.83&64.86&65.00&66.10&\textbf{70.16}&65.23&65.23&65.26&65.26&66.00&67.23&\textbf{70.80}\\
        \bottomrule
    \end{tabular}
    \caption{Comparison of \textit{Misclassification rate} (\%) of IE-Attack and baselines under two detection methods. }
    \label{imperceptibility}
\end{table*}
\textbf{Datasets}
To validate the superiority of our method, five datasets (Cora, Citeseer, Pubmed, Chameleon, Lastfm) \cite{maurya2021improving} are adopted. Following the hypergraph generation methods in the HGNNs, we apply the Hyper-KNN and Hyper-$L$1 methods for hypergraph generation \cite{gao2022hgnn+}. Furthermore, we introduce a novel approach for constructing hypergraphs by considering higher-order neighbors of nodes (Hyper-HOR). Three distinct hypergraph generation strategies are employed to verify the efficacy and robustness of the proposed IE-Attack. According to the datasets partitioning strategy for node classification in Graph Convolutional Networks (GCNs) \cite{kipf2016semi}, datasets are divided into training/validation/test sets. 

\noindent\textbf{Parameter Setting} In this study, we set the elite hyperedge perturbation budget $\eta$, which involves selecting $\eta \times \omega$ hyperedges within elite hyperedge $\mathcal{E}_{elite}$. The value of $\eta$ ranges from 0.1 to 1 and $\omega$ is the number of $\mathcal{E}_{elite}$. Regarding the value of $K$ in the Hyper-KNN hypergraph construction method, we set it to 10. The order in Hyper-HOR is set to 1-order and $\gamma$ in Hyper-$L$1 is set to 0.1. 

\noindent\textbf{Evaluating Metrics} This paper utilizes the \textit{Misclassification rate} as an evaluation metric for the performance of IE-Attack. The \textit{Misclassification rate} indicates the success rate of misclassifications by HGNNs, where a higher \textit{Misclassification rate} signifies a more effective attack.

\noindent\textbf{Baselines}
IE-Attack is the node injection attack against HGNNs. Hence, we modify node injection methods from GNNs to align with the strategy proposed for hypergraphs in this study. We set up six baseline methods, including random methods (GIA-Random (GIA-R), DICE \cite{Waniek:dice2018}), gradient methods (FGA \cite{chen2018fast}, IGA \cite{wu2019adversarial}), and adversarial generation (G-NIA \cite{tao2021single}, G-NIA\_CANA  (G-NIA$^{*}$) \cite{tao2023adversarial}). These baseline methods adapt graph attacks to hypergraphs, making them comparable to proposed IE-Attack. Moreover, IE-Attack and baselines attack HGNNs in evasion attack scenarios. All experiments are conducted on a workstation equipped with four NVIDIA RTX 3090 GPUs, which are conducted under the same parameter settings. Except for Table \ref{performance_all}, the random seed for other experiments is set 2024.
\subsection{Model Performance and Parameter Analysis}
\subsubsection{Performance Comparison with State-of-the-art Methods}
Table \ref{performance_all} displays the \textit{Misclassification rate} of IE-Attack and baseline methods across three hypergraph generation models on five datasets. The ``Clean" shows \textit{Misclassification rate} of the hypergraph without attacks in HGNNs. Except for G-NIA and G-NIA$^{*}$, the other four baseline methods show only slight improvements in attack performance on hypergraphs compared to the ``Clean" results. This suggests that randomly selecting hyperedges and using gradient guidance for node injections are not effective for hypergraph attacks.
In contrast, G-NIA introduces injected nodes into hyperedges generated by a generator after multiple training iterations, leading to improved attack performance. G-NIA$^{*}$ incorporates a discriminator to enhance the imperceptibility of attacks. Nonetheless, IE-Attack achieves the highest \textit{Misclassification rate} across all datasets and hypergraph generation models. These findings show attack effectiveness on HGNNs by injecting the homogeneous node into elite hyperedges.
Moreover, the Cora hypergraph generated by the Hyper-HOR shows higher susceptibility to attacks, indicating weaker robustness. Conversely, the Citeseer hypergraph generated by Hyper-KNN and Hyper-$L$1 is more vulnerable to such attacks. These differences arise from how models capture graph structural information and node features.
\begin{figure}[htbp] 
  \centering  
  \includegraphics[width=0.9\linewidth]{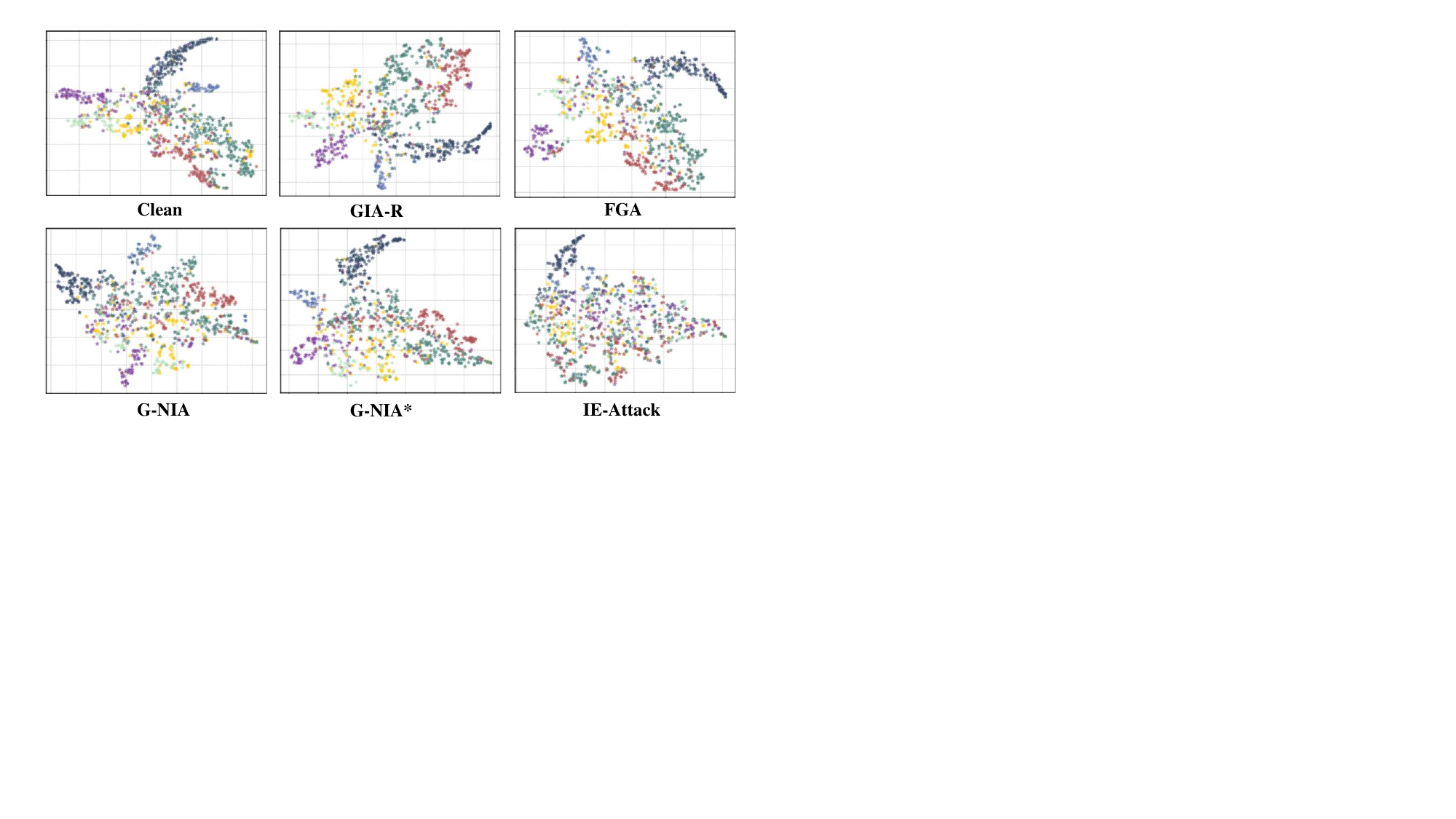}
  \caption{Visualization of node classification obtained by HGNNs after attack on Cora under Hyper-KNN.}
  \label{hyper_class}
\end{figure}

To demonstrate the impact of IE-Attack on HGNNs' classification performance, we visualize the node classification results of Cora under baselines in Figure \ref{hyper_class}. Under the IE-Attack, HGNNs experience a significant decrease in classification performance with substantial data point dispersion.
\subsubsection{Performance Comparison under Detection Methods}
To validate the imperceptibility of IE-Attack, we further attack HGNNs under two detection models (i.e., PCA, HBOS) to obtain attack results, as shown in Table \ref{imperceptibility}. The results demonstrate that even under the detection models, IE-Attack maintains excellent attack performance compared to other baselines. This indicates that proposed node generator with group identity generates homogeneous nodes, while possessing the capability to propagate malicious information. Furthermore, injecting the single homogeneous node into elite hyperedges is also an outstanding strategy to enhance the imperceptibility of attacks.
\subsubsection{Performance of Other Methods for Elite Hyperedges}
\begin{figure}[htbp] 
  \centering
  \includegraphics[width=1\linewidth]{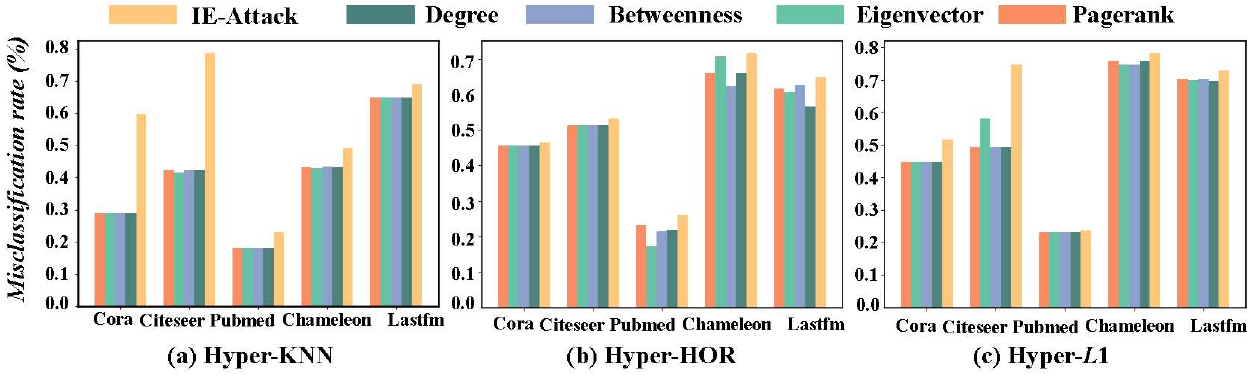}
  \caption{\textit{Misclassification rate} (\%) of IE-Attack compared to other methods for obtaining elite hyperedges.}
  \label{hyperedge_methods}
\end{figure}

To verify that elite nodes identified by the "cycle ratio" effectively enhance attack performance, we compare them with other node importance metrics: Degree centrality, Betweenness centrality, Eigenvector centrality, and PageRank. In Figure \ref{hyperedge_methods}(a), the bar graph illustrates the attack performance using different metrics for the hypergraph generation methods. The height of the bar represents the attack performance of the model caused by the elite hyperedges obtained under this importance metric. The \textit{Misclassification rate} of IE-Attack outperforms the other metrics, suggesting that obtain elite hyperedges by utilizing the node spanning amplify the spread of malicious information.
\subsubsection{Analysis of Perturbation Budget ($\eta$) of Elite Hyperedges}
\begin{figure*}[!htbp]
  \centering
  \includegraphics[width=0.9\linewidth]{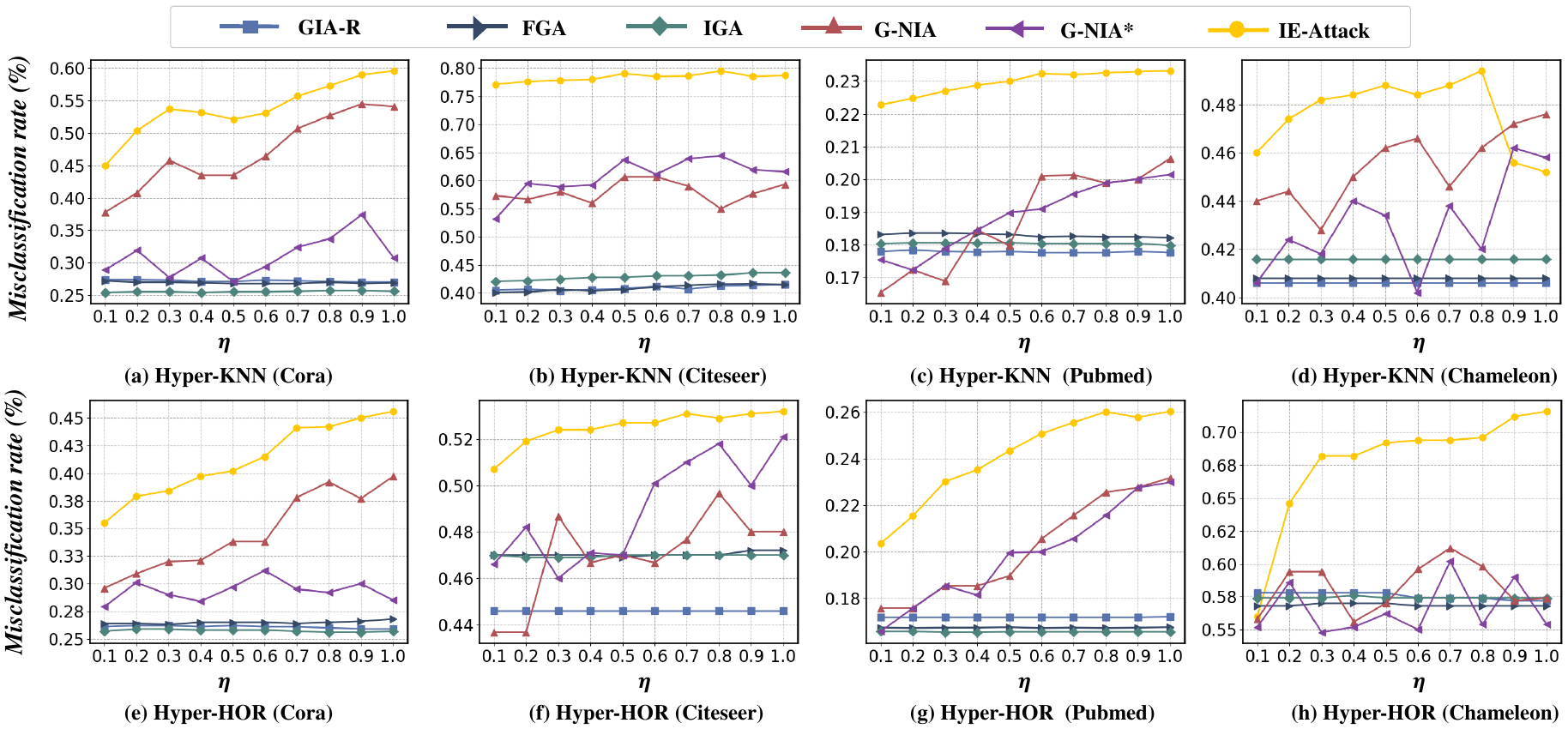}
  \caption{Attack performance of IE-Attack compared to baselines under elite hyperedge perturbation budget $\eta$.}
  \label{hyperedge_buget}
\end{figure*}

Figure \ref{hyperedge_buget} compares the performance of IE-Attack and five baseline methods on four datasets (i.e., Cora, Citeseer, Pubmed, Chameleon) under different hypergraph models as elite hyperedge budget $\eta$ varies from 0.1 to 1. 
For the hypergraphs constructed by Hyper-KNN and Hyper-$L$1, the attack performance of IE-Attack shows an increasing trend across the four datasets as the elite hyperedge budget $\eta$ rises. Surprisingly, GIA-R, FGA, and IGA exhibit minimal changes in \textit{Misclassification rate} as $\eta$ increases, indicating that injecting the node into hyperedges randomly selecting hyperedges and using gradient guidance has little impact on attack performance. Additionally, G-NIA and G-NIA$^{*}$, being node injection attack for graphs applied to hypergraphs, demonstrate lower performance than IE-Attack. 
\subsubsection{Analysis of Kernel Function $K_{de}$}
Table \ref{performance_kernel} illustrates \textit{Misclassification rate} when adopting different kernel functions, i.e., Gaussian($M_{1}$), Tophat($M_{2}$), and Epanechnikov($M_{3}$). Generally, most datasets show optimal attack performance when generating node features using the Gaussian kernel function. Specifically, for the Chameleon, Tophat performs best in Hyper-KNN, while for Lastfm, Epanechnikov demonstrates superior performance in Hyper-HOR. Consequently, the Gaussian kernel function is chosen for generating node features in three citation datasets, while appropriate kernel functions are selected for the remaining two datasets. 
\begin{table}[!htbp]\small
\setlength\tabcolsep{1.5pt}
  \centering
   \begin{threeparttable}
    \begin{tabular}{ll|ccccc} 
    \toprule
    $K_{de}$&Hyper-model&\multicolumn{1}{c}{Cora}&\multicolumn{1}{c}{Citeseer}&\multicolumn{1}{c}{Pubmed}&\multicolumn{1}{c}{Chameleon}&\multicolumn{1}{c}{Lastfm}\cr
     \hline
    \multirow{3}{*}{$M_{1}$} &Hyper-KNN &\textbf{59.60} &\textbf{78.70}&\textbf{23.32}&45.20&\textbf{69.10}\cr
    &Hyper-HOR &\textbf{45.60}&\textbf{53.20}&\textbf{26.04}&\textbf{71.60}&64.70\cr
    &Hyper-$L$1 &\textbf{51.70}&\textbf{74.70}&\textbf{23.66}&\textbf{78.20}&72.87\cr
    \hline
    \multirow{3}{*}{$M_{2}$} &Hyper-KNN&41.60&63.30&22.70&\textbf{46.20}&68.70\cr
    &Hyper-HOR&41.60&52.60&24.50&67.40&64.73\cr
    &Hyper-$L$1 &50.50&55.50&23.34&76.40&73.33\cr
    \hline
    \multirow{3}{*}{$M_{3}$} &Hyper-KNN&38.60&63.90&22.76&45.20&68.87\cr
     &Hyper-HOR&41.60&52.20&24.58&67.20&\textbf{64.77}\cr
     &Hyper-$L$1 &49.70&54.70&23.01&77.80&\textbf{73.37}\cr
    \bottomrule
    \end{tabular}
    \end{threeparttable}
     \caption{Attack performance of different $K_{de}$ in IE-Attack.}
  \label{performance_kernel}
\end{table}
\subsection{Ablation Study and Analysis}
To evaluate the effectiveness of the proposed algorithm, we conduct ablation experiments by systematically removing the ``Elite Hyperedges ($R_{1}$)", ``KDE ($R_{2}$)" and ``Node Generator ($R_{3}$)". Table \ref{performance_ablation} shows that excluding three strategies lead to a decrease in the model's attack performance, resulting in lower \textit{Misclassification rate}. Particularly, the absence of the ``Elite Hyperedges" and ``Node Generator" have a more significant impact on the model's performance. Leveraging the elite hyperedge to maximize the malicious influence of the homogeneous node and utilizing the ``Node Generator" to generate the homogeneous node, IE-Attack enhances its attack performance and improves the imperceptibility of attacks. Ablation experiments confirm the remarkable effectiveness of the proposed IE-Attack on HGNNs.
\begin{table}[htbp]\small
\setlength\tabcolsep{0.6pt}
  \centering
   \begin{threeparttable}
    \begin{tabular}{ll|ccccc}
    \toprule
    Methods&Hyper-model&\multicolumn{1}{c}{Cora}&\multicolumn{1}{c}{Citeseer}&\multicolumn{1}{c}{Pubmed}&\multicolumn{1}{c}{Chameleon}&\multicolumn{1}{c}{Lastfm}\cr
     \hline
    \multirow{3}{*}{w/o $R_{1}$} &Hyper-KNN &30.80&40.70&18.28&43.80&64.83\cr
    &Hyper-HOR &27.10&45.10&16.84&59.00&56.50\cr
    &Hyper-$L$1 &37.30&51.60&23.98&75.40&69.13\cr
    \hline
    \multirow{3}{*}{w/o $R_{2}$} &Hyper-KNN &39.40&65.10&22.72&45.80&68.70\cr
    &Hyper-HOR &42.00&52.20&24.68&71.40&64.23\cr
    &Hyper-$L$1 &50.80&54.90&25.92&78.20&\textbf{73.33}\cr
    \hline
    \multirow{3}{*}{w/o $R_{3}$} &Hyper-KNN &29.50&54.40&19.40&44.40&64.60\cr
     &Hyper-HOR &25.80&48.10&16.88&53.00&55.17\cr
     &Hyper-$L$1 &35.80&48.40&23.10&73.20&70.70\cr
      \hline
    \multirow{3}{*}{IE-Attack} &Hyper-KNN &\textbf{59.60}&\textbf{78.70}&\textbf{23.32}&\textbf{49.40}&\textbf{69.10}\cr
     &Hyper-HOR &\textbf{45.60}&\textbf{53.20}&\textbf{26.04}&\textbf{71.60}&\textbf{64.70}\cr
     &Hyper-$L$1 &\textbf{51.70}&\textbf{74.70}&\textbf{27.02}&\textbf{78.20}&72.87\cr
    \bottomrule
    \end{tabular}
    \end{threeparttable}
     \caption{Attack performance of different IE-Attack variants.}
  \label{performance_ablation}
\end{table}
\section{Conclusion}
This paper explores the phenomenon of node spanning in the hypergraph and the group identity of hyperedges, introducing the node injection attack framework IE-Attack for HGNNs. The key idea lies in injecting the homogeneous node into elite hyperedges, IE-Attack maximizes the spread of malicious information in feature aggregation of HGNNs and enhances the imperceptibility of attacks. 
Extensive experiments demonstrate IE-Attack's superior performance over other attack models. Future research will explore attack techniques for different HGNNs variants and varying levels of knowledge, including black-box attacks.

\section{Acknowledgments}
This work was supported in part by the National Natural Science Foundation of China (Grant nos. 62073263, 62472117); the National Major Research Project (Grant no. 0622-GKGJ30000030094-ZB-Z002-0); the Fundamental Research Funds for the Central Universities (Grant no. D5000230112); and the Open Research Subject of State Key Laboratory of Intelligent Game (Grant no. ZBKF-24-02).

\bibliography{aaai25}

\end{document}